%% file: aaai25.tex
\documentclass[letterpaper]{article} % DO NOT CHANGE THIS
\usepackage{aaai25} %{aaai25}  % DO NOT CHANGE THIS

\usepackage{times}  % DO NOT CHANGE THIS
\usepackage{helvet}  % DO NOT CHANGE THIS
\usepackage{courier}  % DO NOT CHANGE THIS
\usepackage[hyphens]{url}  % DO NOT CHANGE THIS
\usepackage{graphicx} % DO NOT CHANGE THIS
\usepackage{natbib}  % DO NOT CHANGE THIS AND DO NOT ADD ANY OPTIONS TO IT
\usepackage{caption} % DO NOT CHANGE THIS AND DO NOT ADD ANY OPTIONS TO IT
\usepackage{algorithm}
\usepackage{algorithmic}

\usepackage{float}
\usepackage{newfloat}
\usepackage{listings}

\usepackage{booktabs}
\usepackage{multirow}
\usepackage{amssymb}
\usepackage{utfsym}
\usepackage{makecell}
\usepackage{graphicx}
\usepackage{cuted}
\usepackage[most]{tcolorbox}
\tcbuselibrary{fitting}
\DeclareCaptionStyle{ruled}{labelfont=normalfont,labelsep=colon,strut=off} % DO NOT CHANGE THIS
\floatstyle{ruled}
\newfloat{listing}{tb}{lst}{}
\floatname{listing}{Listing}
\nocopyright %-- Your paper will not be published if you use this command
\title{Instruction-guided Multi-Granularity Segmentation and Captioning \\ with Large Multimodal Model}
\author{
    Li Zhou\textsuperscript{\rm 1\equalcontrib},
    Xu Yuan\textsuperscript{\rm 2\equalcontrib},
    Zenghui Sun\textsuperscript{\rm 1},
    Zikun Zhou\textsuperscript{\rm 3},
    Jinsong Lan\textsuperscript{\rm 1}\thanks{Corresponding author.},
}
\affiliations{
    \textsuperscript{\rm 1}TAO Technology, Alibaba Group \\
    \textsuperscript{\rm 2}The Hong Kong Polytechnic University \\
    \textsuperscript{\rm 3}Peng Cheng Laboratory\\
    \{pengye.zl,zenghui.szh,jinsonglan.ljs\}@taobao.com, xuyuan127@gmail.com, zhouzikunhit@gmail.com
}
\usepackage{bibentry}
\begin{document}

\maketitle

% \begin{links}bibliographystyle
    % \link{Code}{https://github.com/lizhou-cs/mglmm}
%     % \link{Datasets}{https://aaai.org/example/datasets}
%     % \link{Extended version}{https://aaai.org/example/extended-version}

\begin{strip}
\begin{minipage}{\textwidth}
\centering
\vspace{-20pt}
\includegraphics[width=\textwidth]{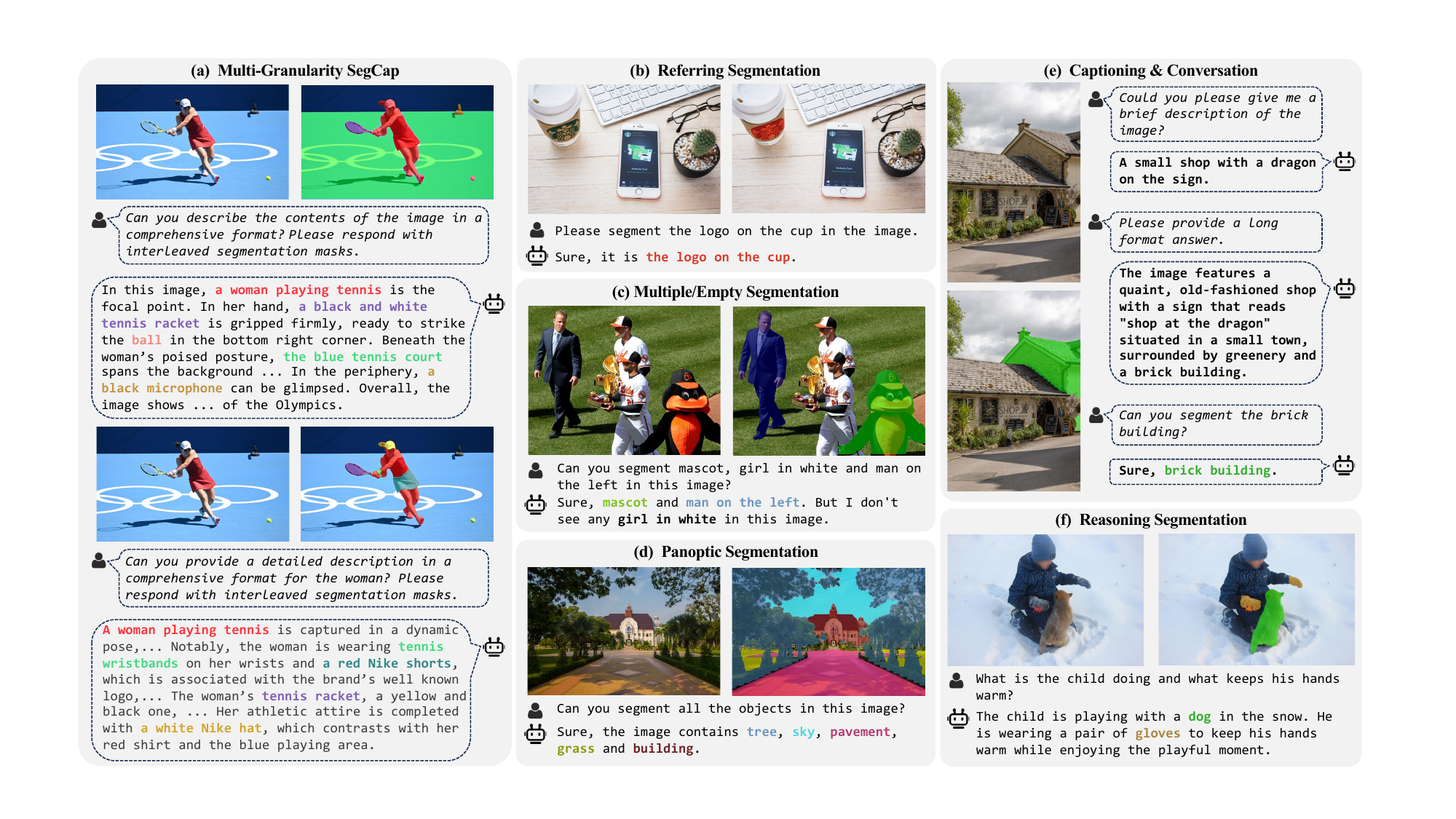}
\captionof{figure}{\small MGLMM is a versatile and sophisticated LMM, which can handle various tasks involving textual and pixel-level mask responses. We show its visualization results in the following scenarios: multi-granularity segmentation and captioning, referring segmentation, multiple/empty segmentation, panoptic segmentation, reasoning segmentation, image-level captioning, and conversation.}
\label{fig:model_capability}
\end{minipage}
\end{strip}
\definecolor{myblue}{RGB}{222,235,247}
% \definecolor{myblue}{RGB}{247,216,180}

\begin{abstract}
\input{1-abstract}
\end{abstract}

%%%%%%%%% BODY TEXT
\section{Introduction}
\input{1-introduction}

\section{Related Work}
\input{2-related_work}

\section{Method}
\input{3-method}

\section{Data Annotation Pipeline}
\input{4-pipeline}

\section{Experiments}
\input{5-experiments}

\section{Conclusion}
\input{6-conclusion}

% \section{Conclusion}

% \bigskip
% \noindent Thank you for reading these instructions carefully. We look forward to receiving your electronic files!

\bibliography{aaai25}

\end{document}

%% file: 1-abstract.tex
Large Multimodal Models (LMMs) have achieved significant progress by extending large language models. 
Building on this progress, the latest developments in LMMs demonstrate the ability to generate dense pixel-wise segmentation through the integration of segmentation models.
Despite the innovations, the textual responses and segmentation masks of existing works remain at the instance level, showing limited ability to perform fine-grained understanding and segmentation even provided with detailed textual cues.
To overcome this limitation, we introduce a Multi-Granularity Large Multimodal Model (MGLMM), which is capable of seamlessly adjusting the granularity of Segmentation and Captioning (SegCap) following user instructions, from panoptic SegCap to fine-grained SegCap. We name such a new task Multi-Granularity Segmentation and Captioning (MGSC). 
Observing the lack of a benchmark for model training and evaluation over the MGSC task, we establish a benchmark with aligned masks and captions in multi-granularity using our customized automated annotation pipeline. 
This benchmark comprises 10K images and more than 30K image-question pairs. 
We will release our dataset along with the implementation of our automated dataset annotation pipeline for further research.
Besides, we propose a novel unified SegCap data format to unify heterogeneous segmentation datasets; it effectively facilitates learning to associate object concepts with visual features during multi-task training. 
Extensive experiments demonstrate that our MGLMM excels at tackling more than eight downstream tasks and achieves state-of-the-art performance in MGSC, GCG, image captioning, referring segmentation, multiple and empty segmentation, and reasoning segmentation tasks. 
The great performance and versatility of MGLMM underscore its potential impact on advancing multimodal research.
Code and dataset will be released at https://github.com/lizhou-cs/mglmm.

%% file: 1-introduction.tex
Leveraging the commonsense reasoning and understanding abilities of Large Language Models (LLMs)~\cite{vicuna2023,touvron2023llama}, Large Multimodal Models (LMMs)~\cite{minigpt4,alayrac2022flamingo,qwenvl,llava} 
have notably advanced cross-modality understanding and vision-language alignment.

\begin{figure}
    \centering
    \includegraphics[width=\linewidth]{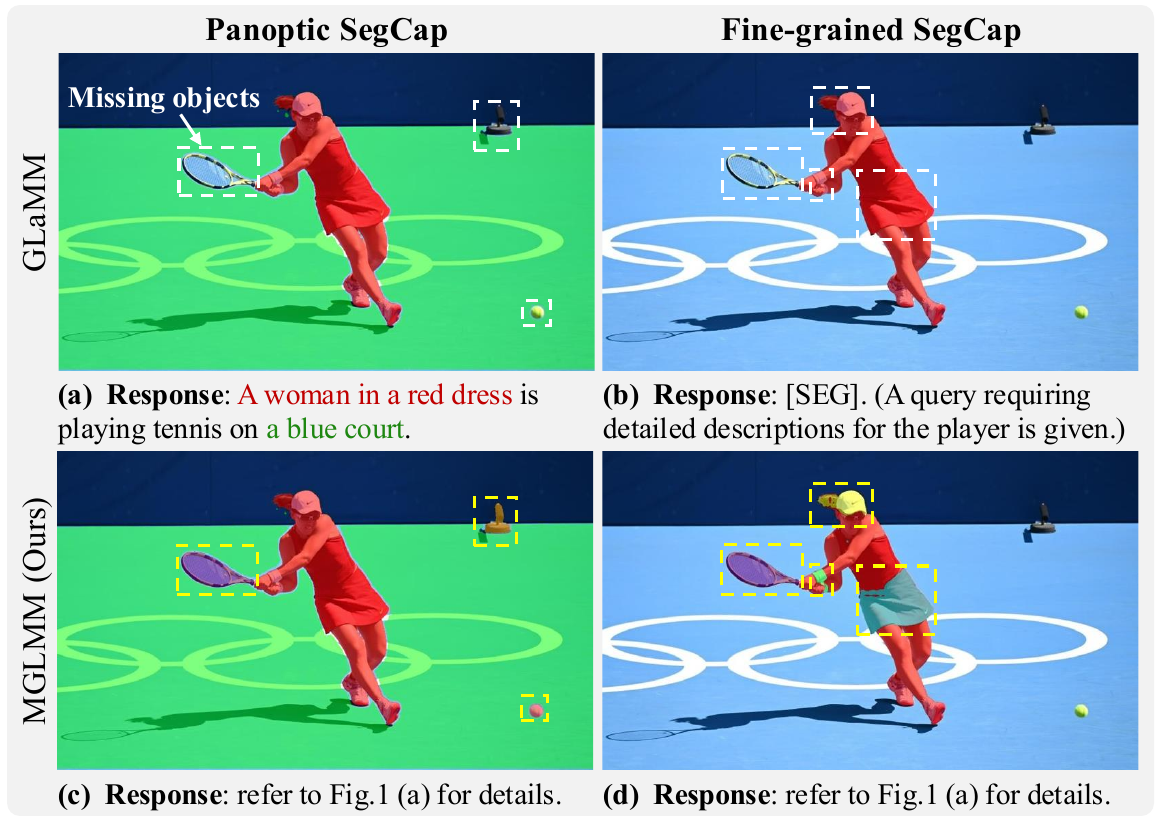}
    \caption{Qualitative comparison of GLaMM and our MGLMM. Please refer to \textbf{Appendix. A} for more details.}
    \label{fig:motivation}
    \vspace{-5pt}
\end{figure}

Recently, several studies ~\cite{lai2024lisa,xia2024gsva} have explored the instruction-based LMMs capable of producing pixel-level segmentation masks as responses to user queries.
More recent researches~\cite{rasheed2024glamm,zhang2024omg} concentrated on Grounded Conversation Generation (GCG) which aims to ground the main objects appearing in the conversations.
Although these methods~\cite{zhang2024omg, lai2024lisa, xia2024gsva, ren2024pixellm} integrate a powerful segmentation model capable of panoptic segmentation, they still have difficulty generating mask-text-aligned responses for all the instances in the image, resulting in limited panoptic segmentation performance. Figure~\ref{fig:motivation} (a) shows such a case where GLaMM overlooks the tennis racket, tennis ball and microphone in both mask and text responses.
Besides, these models only possess the ability to describe the image at the instance level and produce corresponding instance masks aligned with the output texts. 
Hence, these models can hardly perceive the fine-grained objects, such as the hat, wristband, and skirt of the player in Figure~\ref{fig:motivation} (b), even provided with detailed textual cues. 
The missing of the above abilities would limit the universality and comprehension of the LMMs.

To overcome these limitations, we introduce the Multi-Granularity LMM (MGLMM), which is capable of seamlessly adjusting the granularity of Segmentation and Captioning (SegCap) following user instructions, from panoptic SegCap to fine-grained SegCap. To be specific, for the query requiring describing the overall contents of an image, MGLMM outputs the precise panoptic segmentation masks with captions, offering a coarse-grained understanding of the entire image. For the instruction demanding to describe a certain object in the image, MGLMM can produce a detailed response including segmentation masks of the sub-parts of the object as well as corresponding descriptions, which reveal the components of the target object. We name such a task Multi-Granularity SegCap (MGSC), which assesses the ability of progressive cognition from coarse-grained to fine-grained. Overall, MGLMM excels at tackling more than eight downstream tasks such as panoptic SegCap, fine-grained SegCap, GCG, and multiple and empty segmentation, as presented in Figure~\ref{fig:model_capability} and Table~\ref{tab:model-comparison}.

Observing the lack of a benchmark for training and evaluating LMMs for the MGSC task in the community, we establish a new benchmark, dubbed MGSCData, with aligned masks and captions in multi-granularity using the customized automated annotation pipeline. 
It consists of 10K images and over 30K image-question pairs, encompassing both panoptic and fine-grained segmentation. To be more specific, the dataset includes more than 300K segmentation masks, each annotated with a semantic label and an accompanying detailed description. 
MGSCData effectively facilitates the training and assessment of the ability to associate object concepts and visual features in multi-granularity. 
We will release MGSCData and expect it to benefit academia.

Besides the benchmark, another key challenge in unifying segmentation tasks across granularities lies in the significant variation in both the format and semantic level of the queries and outputs. Typically, existing studies directly incorporate the heterogeneous data of different tasks into model training, overlooking the task discrepancies and complicating multimodal alignment further. To handle this issue, we propose a Unified SegCap Data Format (USCDF) to explicitly guide the model in learning the alignment relationships between object concepts and segmentation masks in different granularities during training. Specifically, USCDF unifies the output formats of different segmentation tasks, bridging the gap between them and reducing the difficulty of multi-task learning for the model. The right part of Figure~\ref{fig:framework} illustrates the instantiation of the unified data format on tasks including multi-referring reasoning, panoptic SegCap, and fine-grained SegCap. Experimental results demonstrate that USCDF benefits multi-task learning and vision-language learning. We also evaluate MGLMM across a variety of benchmarks. The experiments demonstrate that it achieves state-of-the-art results on six benchmarks.

In conclusion, our work has four main contributions:
\begin{itemize}
\item We propose MGLMM, the first model capable of seamlessly switching between multi-granularity segmentation and captioning, especially including panoptic and fine-grained segmentation and captioning.
\item We introduce a novel benchmark MGSCData to train and evaluate the ability of multi-granularity segmentation and captioning for LMMs, which comprises over 30K high-quality image-question pairs.
\item We propose a unified data format, which facilitates learning the alignment relationships between object concepts and segmentation masks in multiple granularities.
%, which we believe will advance research in the community.
\item We achieve state-of-the-art performance across various tasks, including MGSC, GCG, image captioning, various segmentation tasks, etc.
\end{itemize}

\begin{table*}[t]
\centering
\resizebox{0.95\linewidth}{!}{
\begin{tabular}{l|cc|ccc|ccc}
    \toprule
    \multirow{2}{*}{Method} 
    & \multicolumn{2}{c|}{Textual Response} 
    & \multicolumn{3}{c|}{Mask Response} 
    & \multicolumn{3}{c}{Textual \& Mask Response} \\
    & Caption & Conversation 
    & Referring Seg & Generic Seg & Multiple/Empty Seg & Reasoning Seg 
    & GCG & MGSC \\
    \midrule
    LISA \cite{lai2024lisa}
    & \checkmark & \checkmark 
    & \checkmark & & & 
    & & \\
    PixelLM \cite{ren2024pixellm}
    & \checkmark & \checkmark 
    & \checkmark & & \checkmark & \checkmark 
    & & \\
    GSVA \cite{xia2024gsva}
    & \checkmark & \checkmark 
    & \checkmark & & \checkmark & 
    & & \\
    Osprey \cite{yuan2024osprey}
    & \checkmark & \checkmark 
    & & & & 
    & & \\
    LaSagnA \cite{wei2024lasagna}
    & & 
    & \checkmark & \checkmark & \checkmark &
    & & \\
    PSALM \cite{zhang2024psalm} 
    & & 
    & \checkmark & \checkmark & \checkmark &
    & & \\
    OMG-LLaVa \cite{zhang2024omg}
    & \checkmark & \checkmark 
    & \checkmark & \checkmark & & 
    & \checkmark & \\
    GLaMM \cite{rasheed2024glamm}
    & \checkmark & \checkmark 
    & \checkmark & & & 
    & \checkmark & \\
    MGLMM (Ours) 
    & \checkmark & \checkmark 
    & \checkmark & \checkmark & \checkmark & \checkmark 
    & \checkmark & \checkmark \\
    \bottomrule
\end{tabular}
}
\caption{\small Comparison of the capabilities of MGLMM with multiple representative methods. Here, ``Generic Seg" comprises semantic segmentation, instance segmentation, and panoptic segmentation; ``Reasoning Seg" requires the model to segment images based on queries involving complex reasoning and provide the corresponding textual interpretation.
 }
\label{tab:model-comparison}
\vspace{-4pt}
\end{table*}

%% file: 2-related_work.tex
Recently, there has been an increasing focus on fine-tuning pre-trained LLMs for visual instructions.
These approaches, including BLIP-2~\cite{li2023blip2}, InstructBLIP~\cite{dai2023instructblip}, LLaVA ~\cite{liu2024llava}, MiniGPT-4~\cite{minigpt4}, Qwen-VL~\cite{qwenvl}, typically employ a pre-trained visual encoder to embedding visual input, utilize an LLM as the base model to comprehend user instructions and generate textual responses, and include an adapter to bridge the features of the vision encoder with those of the language model.
The integration of visual and linguistic modalities within LLMs aims to enhance their capacity to understand and respond to complex, visually guided tasks. 
Although these methods have significantly facilitated the development of multimodal language models, their mechanisms fail to achieve pixel-level alignment and a comprehensive understanding of both images and language.

Furthermore, several works, including \cite{lai2024lisa, ren2024pixellm, rasheed2024glamm, zhang2024omg}, explore more complex tasks driven by instructions, involving segmentation or captioning as responses to achieve effective pixel-level alignment of images and text.
Although these methods perform well in various segmentation tasks, they are limited to learning only instance-level vision-language alignment, preventing them from perceiving fine-grained objects.
Furthermore, all these methods integrate a mask decoder capable of panoptic segmentation into their methods but fail to generate coherent mask-text-align responses, resulting in suboptimal performance.

For the reasons mentioned above, our goal is to develop an LMM that can seamlessly perform panoptic and fine-grained segmentation and captioning based on user instructions.
Further, we establish a high-quality benchmark called MGSC that fills the gap for panoptic and fine-grained segmentation and captioning and introduce our automated annotation pipeline.
Last, we propose a unified data format that facilitates explicit learning of alignment relationships between object concepts and segmentation masks.
MGLMM achieves state-of-the-art performances on over six tasks and ablation results also prove the effectiveness of our methods.

%% file: 3-method.tex
\begin{figure*}
\centering
\includegraphics[width=0.88\linewidth]{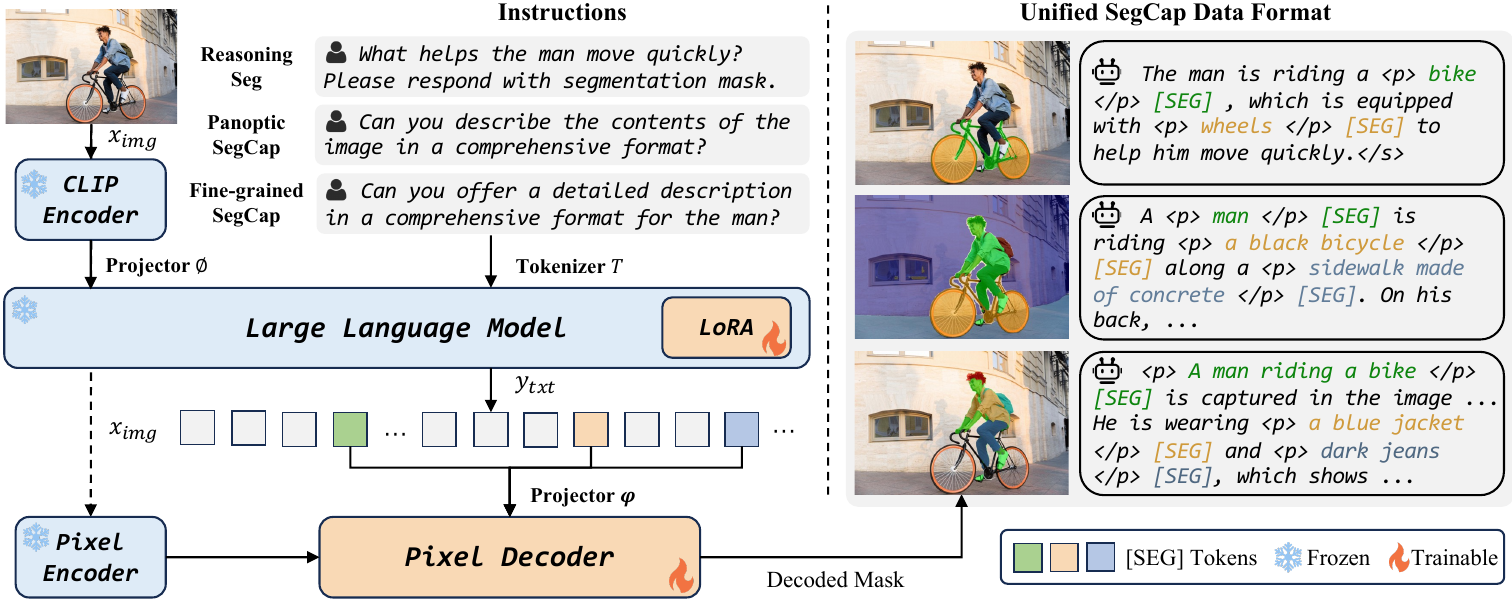}
\caption{\small \textbf{Left}: The model architecture of MGLMM. \textbf{Right}: The proposed unified data format for multi-task learning.}
\label{fig:framework}
\end{figure*}

In this section, we introduce the model architecture of our MGLMM, as illustrated in Figure~\ref{fig:framework}. 
We then introduce the unified SegCap data format used during training.

\vspace{-4pt}
\subsection{Model Architecture}
To achieve multi-granularity segmentation and captioning, we utilize two foundational models to construct our model: 
(1) an LMM for comprehending input images and user instructions and generating natural language responses, and (2) a segmentation model based on an encoder-decoder architecture for pixel-level visual understanding. 

\vspace{1mm}
\noindent\textbf{Large Multimodal Model.}
Considering the simplicity and consistency with previous works \cite{lai2024lisa,rasheed2024glamm}, LLaVA emerges as our preferred choice. 
Specifically, we employ the CLIP model as the vision encoder, denoted as $\mathcal{F}_{v}$, in conjunction with the Vicuna-7B model as a decoder-based LLM, denoted as $\mathcal{F}_{llm}$. 
As illustrated in Figure. \ref{fig:framework}, the vision encoder is responsible for extracting visual features from the input image $x_{img}$, after which a projector  $\phi$ is applied to map the extracted image features into the word embedding space of $\mathcal{F}_{llm}$. 
Formally:
\begin{equation}
    z_{img} = \phi (\mathcal{F}_{v}(x_{img})).
    \label{equ:img_features}
\end{equation}
It is worth noting that the projector $\phi$ plays a crucial role in aligning image features with the linguistic modality. 
Specifically, it consists of two linear layers with a GELU non-linearity and is initialized randomly. 
Meanwhile, the text input is encoded into text tokens by the tokenizer $T$ of $\mathcal{F}_{llm}$. 
Subsequently, we integrate image tokens $z_{img}$ and text tokens $z_{txt}$, which are then fed into the $\mathcal{F}_{llm}$ to generate final textual output $y_{txt}$, \textit{i.e.}, 
\begin{equation}
    \hat{y}_{txt} = \mathcal{F}_{llm} (z_{img} \| z_{txt}).
    \label{equ:pred_text}
\end{equation}

Following LISA \cite{lai2024lisa}, we adopt the embedding-as-mask paradigm to bridge these two modules.
In this paradigm, the vocabulary of the model is augmented with a specialized token `[SEG]', designed to explicitly activate the segmentation behavior of the segmentation model.
When the LMM intends to generate a segmentation mask based on the user instruction, it inserts the `[SEG]' token in the output sequence $y_{txt}$ to indicate the presence of a target to segment. 
For example:
\begin{tcolorbox}[
    % colback=blue!15, %gray background
    colback=myblue, %gray background
    colframe=black,% black frame color
    arc=2pt, auto outer arc,
    width=\linewidth,
    % fontupper=\scriptsize, 
    % fontlower=\large,
    boxrule=0.3mm,
    left=1pt,
    right=1pt,
    top=1pt,
    bottom=1pt]
\textbf{User}: \textless IMAGE\textgreater \ Please segment the dog in this image. \\
\textbf{Assistant}: Sure, the segmentation result is dog [SEG].
\end{tcolorbox}

\vspace{1mm}
\noindent\textbf{Segmentation Model.}
This work employs SAM~\cite{sam} as our foundation segmentation architecture because of its promising pixel-level modeling capability. 
As shown in Figure.~\ref{fig:framework}, the pixel encoder $\mathcal{E}_{pixel}$ is instantiated using a frozen SAM encoder, while the pixel decoder $\mathcal{D}_{pixel}$ is initialized from the pre-trained SAM decoder. 
The former takes the high-resolution image as input to extract fine-grained visual information, while the latter generates the desired segmentation masks prompted by the embedding of the `[SEG]' token from the LLM. 
Specifically, we select the output embedding $\hat{z}_{seg}$ corresponding to the `[SEG]' token $\hat{y}_{txt}([SEG])$ and transform it into the feature space of decoder using a projector $\psi$. 
Notably, the structure and initialization of projector $\psi$ are identical to those of projector $\phi$. 
The entire process can be formulated as:
\begin{equation}
    \hat{y}_{mask} = \mathcal{D}_{pixel} (\mathcal{E}_{pixel}(x_{img}), \psi(\hat{z}_{seg})).
    \label{equ:pred_mask}
\end{equation}

\begin{figure*}[t]
\centering
\includegraphics[width=\linewidth]{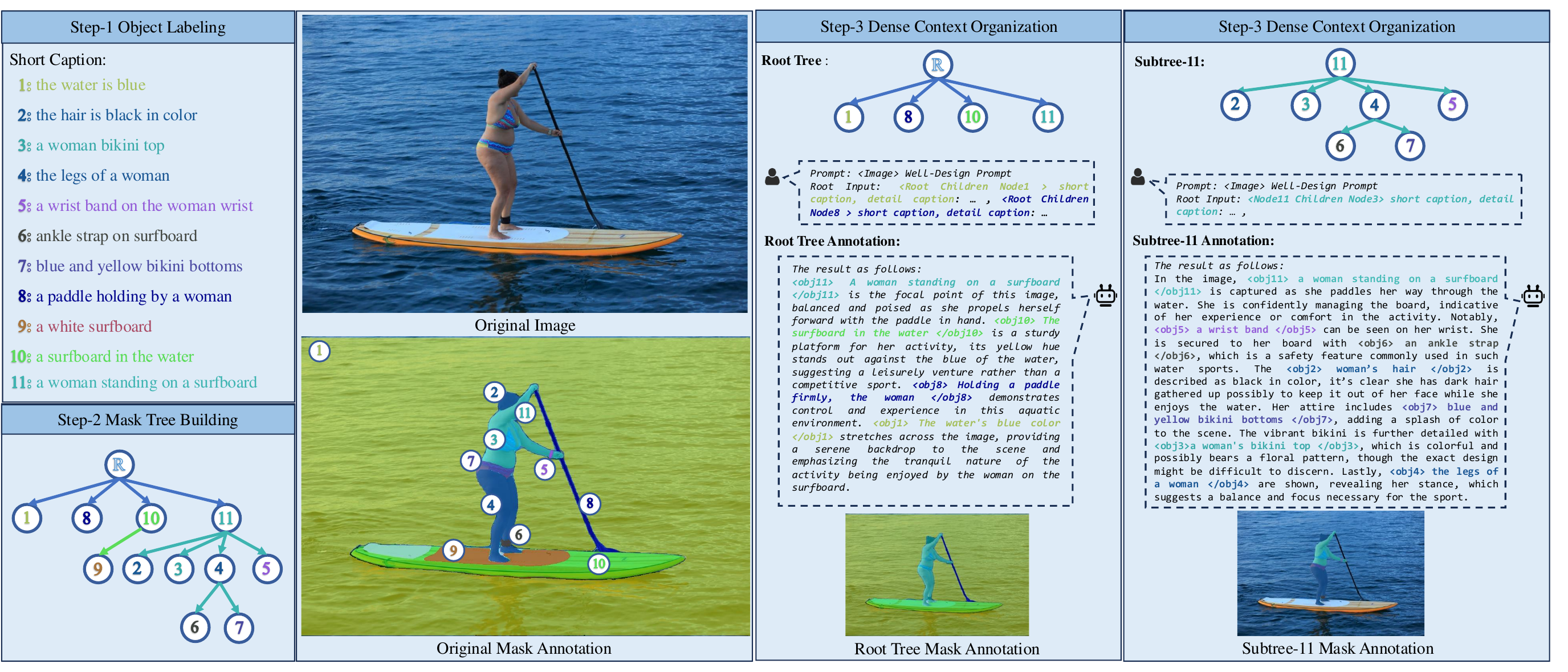}
\caption{\small The overview of our proposed data auto-annotated pipeline. 
Due to space limitations, the detailed caption is not shown in the figure.
Please refer to the \textbf{Appendix. B} for the detailed version. 
Best viewed with zoom-in.}
\label{fig:pipeline}
\vspace{-5pt}
\end{figure*}

\subsection{Design of Unified SegCap Data Format}
\label{sec:unified-data-schema}
Most existing studies primarily integrate various pixel-level segmentation capabilities into LMMs by directly extending corresponding task datasets.
For example, in referring segmentation, the query may be a phrase that requires the return of segmentation masks. 
Conversely, in reasoning segmentation, the query can be a longer sentence or question in which the target may not be present, necessitating an answer along with segmentation masks.
In different segmentation tasks, the form and semantics of queries vary. 
In this context, the model must adaptively align the semantic concepts of potential targets with visual features during training, which undoubtedly increases the burden on model learning.
Therefore, we propose a unified SegCap data format to leverage these data, explicitly guiding the model toward improved vision-language alignment. 
In this manner, we unify the output formats of different segmentation tasks, bridging the gap between them and reducing the difficulty of multi-task learning for the model.
Specifically, apart from the `[SEG]' token, we also introduce \textless p\textgreater \ and \textless /p\textgreater \ tokens to the vocabulary of the LMM to denote the start and end of the corresponding phrases of the segmentation mask, respectively. 
The LLM is required to mark the corresponding description with \textless p\textgreater \ and \textless /p\textgreater \ while activating the segmentation behavior using `[SEG]'. The following is an example of data format for multi-referring segmentation:
\begin{tcolorbox}[
    colback=myblue, %gray background
    colframe=black,% black frame colour
    arc=2pt, auto outer arc,
    width=\linewidth,
    boxrule=0.3mm,
    % fontupper=\slshape,
    % fontupper=\scriptsize, 
    % fontlower=\large,
    left=1pt,
    right=1pt,
    top=1pt,
    bottom=1pt]
\textbf{User}: \textless IMAGE\textgreater \ Please segment the \{obj-1\}, \{obj-2\}, ..., and \{obj-n\} in this image. \\
\textbf{Assistant}: Sure, 
\textless p\textgreater \ \{obj-1\} \textless /p\textgreater \ [SEG], 
\textless p\textgreater \ \{obj-2\} \textless /p\textgreater \ [SEG], ..., and 
\textless p\textgreater \ \{obj-n\} \textless /p\textgreater \ [SEG].
\end{tcolorbox}

\noindent Here, \textless IMAGE\textgreater \ denotes the placeholder for image tokens. \{obj-n\} represents the semantic description of the corresponding segmentation targets.

In contrast to previous work, such a unified data pattern enables the model to explicitly learn the alignment relationships between the object concepts and the segmentation masks during training. 
Despite the fact that GLaMM \cite{rasheed2024glamm} had adopted a similar format, it was only employed for the GCG task it presented. 
In contrast, we utilize this unified schema for all tasks, which reduces the modeling burden by minimizing the differences in output formats across tasks. 
In Figure. \ref{fig:framework}, we demonstrate our unified data format on tasks such as reasoning and multi-granularity segmentation.
Notably, during the training phase, we convert the annotation format of some existing open-source datasets into the proposed unified data schema as they do not meet our requirements. For more details on this process, please refer to \textbf{Appendix. C}.

%% file: 4-pipeline.tex
Most existing segmentation datasets focus on instance-level objects, and although the SAM dataset provides fine-grained segmentation mask annotations, it lacks corresponding text descriptions.
Therefore, to address the issue of insufficient benchmarks for evaluating models in multi-granular segmentation and captioning, we propose a novel task called Multi-Granularity SegCap.
To build up this benchmark, we came up with an automated annotation pipeline that allows us to leverage the capabilities of LMMs, specifically the GPT-4 and Qwen-VL series, for data labeling.
In the following section, we will introduce our automatic annotation pipeline, designed to seamlessly transform any segmentation dataset. 
This pipeline consists of three main steps, as illustrated in Figure~\ref{fig:pipeline}. 
The first step focuses on generating short captions and detailed captions for each masked target, known as object labeling. 
Subsequently, the second step constructs tree relationships based on the segmentation masks. 
The third step organizes various levels of granular information by utilizing the raw data from different levels of the subtree.
As a result, we achieve multi-granularity segmentation and captioning annotations that demonstrate high alignment between visual and textual concepts.
Since the SAM \cite{sam} dataset provides hundreds of millions of high-quality images and fine-grain segmentation, we perform our automated pipeline on the SAM dataset.

\subsection{Object Labeling}
In step 1, the key point is generating a short caption and detailed caption for each target in the images. 
The short caption is used as a semantic representation of the target.
The detailed caption is a comprehensive and semantically rich textual representation of the target, which is primarily used to provide a reference representation to limit the divergence and randomness of LMMs.
In practice, we leverage the GPT-4o to create instruction-following data to generate the semantic label of each masked object.

\subsection{Mask Tree Building}
After obtaining the semantic labels of each target, we need to organize the hierarchical relationships between each target within the image.
We discover that the hierarchical relationships between the targets could be effectively reflected by the Intersection of Union (IoU) relationships among the masks.
Therefore, we denote the entire image as the root node and then extend the tree according to the inclusion relationship between masks.
Besides, in the SAM dataset, numerous mask annotations exist within a single image, many of which share the same semantics labels. 
For example, in a building with many windows, each window is represented as an individual mask with the same short captions. 
For such nodes that share the same parent node, we merge the nodes and their masks.
In this manner, we obtain a simple and hierarchical tree and significantly shorten the length of the resulting text annotations.

\begin{table*}[!tbp]
\centering
\resizebox{\linewidth}{!}{
\begin{tabular}{l|cc|cccc|cccc}
\toprule
  \multirow{3}{*}{Method} &
  \multicolumn{2}{c|}{Textual Response} &
  \multicolumn{4}{c|}{Mask Response} &
  \multicolumn{4}{c}{Textual \& Mask Response} \\
  \cmidrule{2-3}
  \cmidrule{4-7}
  \cmidrule{8-11}
  & Flickr30k &
  NoCap &
  refCOCO+ &
  refCOCOg &
  gRefCOCO &
  reasonSeg &
  \multicolumn{2}{c}{GCG} &
  \multicolumn{2}{c}{MGSC} \\
          & CIDEr & CIDEr & cIoU & cIoU & cIoU & cIoU & CIDEr & AP50 & CIDEr & AP50 \\
\midrule
LISA \cite{lai2024lisa}       & --     & --     & 65.1 & 67.9 & --    & 46.0 & 33.9  & 25.2 & --     & --    \\
PixelLM \cite{ren2024pixellm} & --     & --     & 66.3 & 69.3 & --    & --    & --     & --    & --     & --    \\
GSVA \cite{xia2024gsva}       & --     & --     & 65.9 & 72.7 & --    & --    & --     & --    & --     & --    \\
LaSagnA \cite{wei2024lasagna} & --     & --     & 66.4 & 70.6 & 38.1 & 47.2 & --     & --    & --     & --    \\
PSALM \cite{zhang2024psalm}   & --     & --     & 72.9 & 73.8 & 42.0 & --    & --     & --    & --     & --    \\
OMG-LLaVA \cite{zhang2024omg} & --     & --     & 69.1 & 72.9 & --    & --    & 41.2  & 29.9 & --     & --    \\
GLaMM \cite{rasheed2024glamm} & 95.3  & 106.8 & 72.6 & 74.2 & --    & --    & 47.2  & 30.8 & 8.7   & 5.4  \\
\midrule
MGLMM (Ours)&
  \textbf{104.6} &
  \textbf{112.6} &
  \textbf{73.9} &
  \textbf{77.2} &
  \textbf{52.8} &
  \textbf{51.1} &
  \textbf{50.1} &
  \textbf{31.7} &
  \textbf{11.6} &
  \textbf{7.4} \\
\bottomrule
\end{tabular}
}
\caption{The comprehensive comparison of MGLMM and other LMMs in terms of text description and pixel-level understanding capabilities. ``--" indicates that the method does not handle this task.}
\vspace{-5pt}
\label{tab:overview}
\end{table*}

\subsection{Dense Context Organization}
The generation of multi-granularity captions is based on the mask tree which provides semantic labels of each target and hierarchical relationships between them.
First, we utilize the semantic labels of child nodes of the root node to generate an ordered text input which mainly includes the instance-level objects in the image, which aims to create a coarse-grained caption for the entire picture.
Subsequently, we concatenate the well-designed prompt, the ordered text input, and the image to prompt GPT-4o and obtain an organized description in which each target is embedded in a natural and coherent sequence.
We apply the same process on each subtree under the root node.
In particular, we use all the descendant nodes of the subtree to build up a description aiming to obtain a fine-grained description of the specific target.
Through such a construction process, we obtain panoptic segmentation masks with aligned descriptions for each instance-level target, as well as fine-grained segmentation masks with aligned descriptions for the specific target in each image.

In this manner, we annotate 10K SAM images, which are inherently diverse and exhibit multi-granularity. 
The resulting dataset comprises 30K conversations and contains over 45M tokens, totaling more than 300K segmentation masks, each accompanied by a short semantic label and a detailed caption.
For more details about the pipeline and dataset, please refer to the \textbf{Appendix. B}.

%% file: 5-experiments.tex
\subsection{Experimental Settings}
\vspace{1mm}
\noindent\textbf{Datasets.}
To achieve all the capabilities of MGLMM, our training dataset is composed of six parts: 
(1) semantic segmentation: including ADE20K \cite{zhou2019ade20k}, COCO-Stuff \cite{caesar2018cocostuff}, Maplilary Vistas \cite{neuhold2017mapillary}, PACO-LVIS \cite{ramanathan2023paco}, and PASCAL-Part \cite{chen2014pascalpart} ;
(2) referring segmentation: including RefCLEF \cite{jing2021refclef} the RefCOCO series \cite{yu2016refcocos}; 
(3) image-level caption: including COCO Caption \cite{chen2015coco}; 
(4) visual question answering: including LLaVA-150k \cite{liu2024llava}
(5) grounded conversation generation including GranDf. Additionally, we also use approximately 4M captioning and referring segmentation data from Grounding-anything Dataset (GranD) \footnote{Although GranD contains 11M images, only 4M are available because the authors have yet to publicize all the data.} dataset published by GLaMM \cite{rasheed2024glamm}, which is annotated automatically on SAM \cite{sam} images.
(6) multi-granularity SegCap, including MGSCData, which we proposed.

\vspace{1mm}
\noindent\textbf{Implementation details.}
In our experiments, we use Vicuna-7B as a structure for LLM except for some ablations. 
We train our model on 16 Tesla A100 GPUs (80GB) for 30,000 iterations with a batch size of 16 per device. 
Unless otherwise specified, the model is trained with a joint training setting and without additional task-specific fine-tuning. 
Following the previous works, we apply the CE loss for modeling text generation, and the BCE and DICE loss to supervise high-quality mask prediction.
Further implementation details, particularly regarding LORA fine-tuning, the optimizer, hyperparameter settings, and training objectives, can be found in the \textbf{Appendix. D}.

\begin{table}[htbp]
\setlength\tabcolsep{3pt}
\resizebox{\linewidth}{!}{
\begin{tabular}{l|ccccc|ccccc}
\toprule
\multirow{2}{*}{Model} & \multicolumn{5}{c|}{Multi-Granularity SepCap} & \multicolumn{5}{c}{GCG}       \\
                       & M    & C    & AP50 & mIoU & MR & M    & C    & AP50 & mIoU & MR \\
\midrule
Kosmos-2               & --   & --   & --   & --   & --   & 16.1 & 27.6 & 17.1 & 55.6 & 28.3   \\
LISA                   & --   & --   & --   & --   & --   & 13.0 & 33.9 & 25.2 & 62.0 & 36.3   \\
OMG-LLaVA              & --   & --   & --   & --   & --   & 14.9 & 41.2 & 29.9 & 65.5 & --     \\
GLaMM                  & 16.5 & 8.7  & 5.4  & 47.6 & 18.7 & 16.2 & 47.2 & 30.8 & \textbf{66.3} & 41.8          \\
\midrule
MGLMM (Ours) 
& \textbf{17.8} & \textbf{11.6} & \textbf{7.4}  & \textbf{51.6} & \textbf{23.2} 
& \textbf{16.4} & \textbf{50.1} & \textbf{31.7} & \textbf{66.3} & \textbf{45.2} \\
\bottomrule
\end{tabular}
}
\caption{\small Performance comparison on MGSC and GCG. Following the evaluation protocol of GCG, we report the metrics including METEOR (M), CIDEr (C), AP50, mIoU, and Mask Recall (MR). 
% metrics.
\vspace{-5pt}
}
\label{tab:gcg}
\end{table}

\vspace{1mm}
\noindent\textbf{Comparisons with State-of-the-Arts.}
As shown in Table~\ref{tab:overview}, we compare our MGLMM with other representative methods on various kinds of tasks and outperform all tasks.
Then, we evaluate the effectiveness of our MGLMM on the following six benchmarks.
Additionally, we will provide more discussion of the experimental results in the Appendix. E.

\vspace{1mm}
\noindent\textbf{Multi-Granularity SegCap.}
The MGSC aims to evaluate the ability to seamlessly adjust the granularity of segmentation and captioning.
Following the same settings, we finetune the GLaMM and our MGLMM on the training set of MGSCData and evaluate them on the same metric.
As shown in Table~\ref{tab:gcg}, we outperform GLaMM on every metric, demonstrating the impressive capabilities of our MGLMM in multi-granularity SegCap.

\vspace{1mm}
\noindent\textbf{Grounded Conversation Generation (GCG).}
Following GLaMM, we finetune our model on the GranDf dataset. 
As shown in Table~\ref{tab:gcg}, our MGLMM outperforms other approaches in terms of both image description and pixel understanding capabilities.
It is worth noting that, despite more training data utilized by GLaMM in the pre-training phase compared to MGLMM, the latter still surpasses the former, particularly in terms of the CIDEr and Mask Recall scores.

\begin{table*}[!tbp]
\centering
\resizebox{.85\linewidth}{!}{
\begin{tabular}{c|l|ccc|ccc|cc|cc}
\toprule
\multirow{2}{*}{Type} 
& \multirow{2}{*}{Model} 
& \multicolumn{3}{c|}{refCOCO}                   
& \multicolumn{3}{c|}{refCOCO+}                  
& \multicolumn{2}{c|}{refCOCOg}  
& \multicolumn{2}{c}{ReasonSeg}
\\
            & & val  & testA & testB & val  & testA & testB & val  & test & cIoU & gIoU \\
\midrule
\multirow{3}{*}{\makecell[c]{Segmentation \\ Specialist}}
& LAVT \cite{yang2022lavt}            & 72.7 & 75.8 & 68.8 & 62.1 & 68.4 & 55.1 & 61.2 & 62.1 & -- & -- \\
& ReLA \cite{liu2023gres}             & 73.8 & 76.5 & 70.2 & 66.0 & 71.0 & 57.7 & 65.0 & 66.0 & -- & -- \\
& PolyFormer \cite{liu2023polyformer} & 74.8 & 76.6 & 71.1 & 67.6 & 72.9 & 59.3 & 67.8 & 69.1 & -- & -- \\
\midrule
\multirow{8}{*}{\makecell[c]{LMM-based \\ Models}}

& LISA \cite{lai2024lisa}  & 74.9 & 79.1  & 72.3  & 65.1 & 70.8  & 58.1  & 67.9 & 70.6 & 46.0 & 34.1  \\
& PixelLM \cite{ren2024pixellm}   & 73.0 & 76.5  & 68.2  & 66.3 & 71.7  & 58.3  & 69.3 & 70.5 &  --   &  --   \\
& GSVA \cite{xia2024gsva}     & 77.2 & 78.9  & 73.5  & 65.9 & 69.6  & 59.8  & 72.7 & 73.3 &  --   &  --   \\
& LaSagnA \cite{wei2024lasagna}   & 76.8 & 78.7  & 73.8  & 66.4 & 70.6  & 60.1  & 70.6 & 71.9 & 47.2 &  --   \\
& OMG-LLaVA \cite{zhang2024omg}   & 78.0 & 80.3  & 74.1  & 69.1 & 73.1  & 63.0  & 72.9 & 72.9 &  --   &  --   \\
& GLaMM \cite{rasheed2024glamm}   & 79.5 & 83.2  & 76.9  & 72.6 & 78.7  & 64.6  & 74.2 & 74.9 &  --   &  --   \\
\cmidrule{2-12}
& MGLMM (Ours)\dag & 80.2 & 83.1 & 76.0 & 73.2 & 78.7 & 66.8 & 76.7 & \textbf{77.5} & \textbf{51.1} &  \textbf{48.6} \\
& MGLMM (Ours)& \textbf{81.3} & \textbf{83.5} & \textbf{77.3} & \textbf{73.9} & \textbf{79.2} & \textbf{67.2} & \textbf{77.2} & 77.4 & -- & -- \\
\bottomrule
\end{tabular}
}
\caption{\small Performance on referring and reasoning segmentation benchmarks. The table only shows the cIoU values for referring segmentation. MGLMM\dag \ indicates that the referring segmentation dataset is used only in the pre-training phase.}
\vspace{-10pt}
\label{tab:ref-seg}
\end{table*}

\vspace{1mm}
\noindent\textbf{Referring Segmentation.}
Table~\ref{tab:ref-seg} compares our MGLMM with current state-of-the-art models on three representative datasets. 
We achieve significant lead performances over recent works like GLaMM, and OMG-LLaVG on the refCOCO/+/g validation and test sets in Table~\ref{tab:ref-seg}.
Notably, even without any fine-tuning on the referring segmentation dataset (MGLMM\dag \ in Table~\ref{tab:ref-seg}), our approach still surpasses GLaMM on the validation split of all benchmarks.

\begin{table}[t]
\setlength\tabcolsep{4pt}
\centering
\resizebox{\linewidth}{!}{
\begin{tabular}{l|c|cccccc}
\toprule
\multirow{3}{*}{Model} & \multirow{3}{*}{\makecell[c]{zero \\ shot}} & \multicolumn{6}{c}{Generalized Referring Segmentation} \\
& &
\multicolumn{2}{c}{val} &
\multicolumn{2}{c}{testA} &
\multicolumn{2}{c}{testB} \\
&  & cIoU & gIoU & cIoU & gIoU & cIoU & gIoU \\
\midrule
% MattNet \cite{yu2018mattnet}  & \usym{2717} & 47.5 & 48.2 & 58.7 & 59.3 & 45.3 & 46.1 \\
% VLT \cite{ding2021vlt}        & \usym{2717} & 52.3 & 52.7 & 61.9 & 62.6 & 49.9 & 50.4 \\
ReLA \cite{liu2023gres}       & \usym{2717} & 62.4 & 63.6 & 69.3 & 70.0 & 59.9 & 61.0 \\
LISA\dag \cite{lai2024lisa}   & \usym{2717} & 38.7 & 32.2 & 52.6 & 48.5 & 44.8 & 39.7 \\
LISA \cite{lai2024lisa}       & \usym{2717} & 61.7 & 61.6 & 69.2 & 70.1 & 60.3 & 61.3 \\
GSVA\dag \cite{xia2024gsva}   & \usym{2717} & 61.7 & 63.3 & 69.2 & 70.1 & 60.3 & 61.3 \\
GSVA \cite{xia2024gsva}       & \usym{2717} & 63.3 & 66.5 & 69.9 & 71.1 & 60.5 & 62.2 \\
\midrule
LaSagnA \cite{wei2024lasagna} & \usym{2713} & 38.1 & 32.4 & 50.4 & 47.3 & 42.1 & 38.9 \\
PSALM \cite{zhang2024psalm}   & \usym{2713} & 42.0 & 43.3 & 52.4 & 54.5 & 50.6 & 52.5 \\
MGLMM (Ours) & \usym{2713} &
  \textbf{52.8} &
  \textbf{50.2} &
  \textbf{61.2} &
  \textbf{58.7} &
  \textbf{56.0} &
  \textbf{54.1} \\
\bottomrule
\end{tabular}
}
\caption{\small Performance comparison on generalized referring-expression segmentation with cIoU and gIoU metrics. 
LISA\dag\ and GSVA\dag\ exclusively use the gRefCOCO dataset during the pre-training phase, while MGLLM performs zero-shot learning.
}
\label{tab:gref-seg}
\end{table}

\vspace{1mm}
\noindent\textbf{Generalized Referring Segmentation and Reasoning Segmentation.}
The results are shown in Table~\ref{tab:gref-seg}. 
Compared with PSLAM \cite{zhang2024psalm}, the state-of-the-art method in the zero-shot setting, our MGLMM accomplishes average boosts of 6.0\% and 6.5\% in terms of cIoU and gIoU, respectively. Notably, MGLMM even outperforms LISA\dag \ in all cases, which incorporate gRefCOCO during the pre-training phase. For reasoning segmentation, we utilize the validation set of ReasonSeg dataset \cite{lai2024lisa} as the benchmark. 
From the results reported in Table~\ref{tab:ref-seg}, we can observe that the reasoning proficiency of MGLMM surpasses that of other methods.

\begin{table}[t]
\centering
\resizebox{\linewidth}{!}{
\begin{tabular}{l|cc|cc}
\toprule
\multirow{2}{*}{Model} & \multicolumn{2}{c|}{Flickr30k}  & \multicolumn{2}{c}{NoCap}      \\
                       & CIDEr          & SPICE         & CIDEr          & SPICE         \\
\midrule
% VinVLM \cite{zhang2021vinvl}            & -              & -             & 95.5           & 13.5          \\
% SimVLM \cite{wang2021simvlm}            & -              & -             & 110.3          & 14.5          \\
LEMON \cite{hu2022lemon}                & --              & --             & 106.8          & 14.1          \\
CoCa \cite{yu2022coca}                  & --              & --             & 120.6          & 15.5          \\
% BLIP \cite{li2022blip}                  & -              & -             & 113.2          & 14.7          \\
BLIP-2 \cite{li2023blip2}               & --              & --             & 121.6          & \textbf{15.8} \\
InstructBLIP \cite{dai2023instructblip} & 82.8           & --             & \textbf{123.1} & --             \\
\midrule
Kosmos-1 \cite{huang2024kosmos1}        & 67.1           & 14.5          & --              & --             \\
Kosmos-2 \cite{peng2023kosmos2}         & 66.7           & --             & --              & --             \\
GLaMM \cite{rasheed2024glamm}           & 95.3           & 18.8          & 106.8          & \textbf{15.8} \\
% \midrule
MGLMM (Ours)& \textbf{104.6} & \textbf{22.7} & 112.6          & 15.2 \\
\bottomrule
\end{tabular}
}
\caption{\small Performance comparison on image-level captioning.}
\label{tab:caption}
\end{table}

\vspace{1mm}
\noindent\textbf{Image-level Captioning.}
To investigate this capability, we finetune MGLMM on the Flickr-30K \cite{plummer2015flickr30k} and evaluate Flickr-30K and NoCap \cite{agrawal2019nocaps}, where the latter can be considered as a \textbf{zero-shot} scene. As reported in Table~\ref{tab:caption}, 
MGLMM is superior to the counterpart model GLaMM on several metrics. 

\begin{table}[htbp]
\resizebox{\linewidth}{!}{
\begin{tabular}{l|cc|ccc|cc}
\toprule
\multirow{2}{*}{Model} & 
\multirow{2}{*}{\makecell[c]{ + USCDF}} &  
\multirow{2}{*}{\makecell[c]{ + GranD\\ Dataset}} &
\multicolumn{3}{c|}{refCOCO+} & 
\multicolumn{2}{c}{GCG} \\
& & & val & testA & testB & C & mIoU \\
\midrule
% OMG-LLaVA &             &             & 69.1 & 73.1 & 63.0 & 41.2 & 65.5 \\
% GLaMM     &             &             & 72.6 & 78.7 & 64.6 & 47.2 & 66.3 \\
% \midrule
MGLMM-7B  &             &             & 67.2 & 74.1 & 58.9 & 46.5 & 65.3 \\
MGLMM-7B  & \usym{2713} &             & 69.9 & 76.2 & 62.5 & 46.3 & 65.6 \\
MGLMM-7B  &             & \usym{2713} & 71.4 & 76.9 & 64.0 & 48.0 & 66.2 \\
MGLMM-7B  & \usym{2713} & \usym{2713} & 73.2 & 78.7 & 66.8 & 50.1 & 66.3 \\
MGLMM-13B & \usym{2713} & \usym{2713} 
& \textbf{73.4} 
& \textbf{79.8} 
& \textbf{68.0} 
& \textbf{50.5} 
& \textbf{66.4} \\
\bottomrule
\end{tabular}
}
\caption{\small Ablation study results. For refCOCO+, we utilize cIoU as the metric. `C' denotes the CIDEr score. We implement MGLMM-13B using Llama2-13B as the structure for LLM.}
% \vspace{-10pt}
\label{tab:ablation}
\end{table}

\vspace{-5pt}
\subsection{Ablation Studies}
To perform a thorough ablation study, we assess different variants of MGLMM using two representative benchmarks, \textit{i.e.}, referring segmentation and GCG, which can demonstrate the models' ability to understand pixel-level details and provide image descriptions. For more details, please refer to \textbf{Appendix. E}.

\vspace{1mm}
\noindent\textbf{Effectiveness of USCDF.}
Compared to the 1st variant in Table~\ref{tab:ablation}, MGLMM using USCDF obtains an improvement of more than 2\% on challenging regCOCO+ benchmark.
The performance difference between the 3rd and 4th variants is significant, as GranD is four times larger than the other pre-training data, which further amplifies the gains of USCDF.

\vspace{1mm}
\noindent\textbf{Influence of GranD dataset.}
To investigate the impact of the extra GranD dataset on MGLMM, we experiment without 4M GranD samples. 
Comparing the 2nd and 4th variants in Table~\ref{tab:ablation}, we can find that the GranD dataset contributes a gain.
Despite not utilizing GranD, our MGLMM remains superior to models such as OMG-LLAVA in most cases, ranking second only to GLaMM, which employed over ten times training data during the pre-training phase.

%% file: 6-conclusion.tex
We propose MGLMM, the first model capable of seamlessly adjusting the granularity of segmentation and captioning following user instructions.
Realizing the lack of multi-granularity of segmentation and captioning dataset and benchmark, we introduce a novel benchmark MGSCData to train and evaluate the ability of multi-granularity segmentation and captioning for LMMs, which comprises over 30K high-quality image-question pairs.
To facilitate aligning object concepts with visual features during various segmentation tasks, we propose a unified data format.
Our model excels at tackling more than eight downstream
tasks and outperforms various benchmarks.